\documentclass{article}

\PassOptionsToPackage{numbers, compress}{nonatbib}
\usepackage[nonatbib, final]{nips_2018}

\usepackage[utf8]{inputenc} % allow utf-8 input
\usepackage[T1]{fontenc}    % use 8-bit T1 fonts
\usepackage{hyperref}       % hyperlinks
\usepackage{url}            % simple URL typesetting
\usepackage{booktabs}       % professional-quality tables
\usepackage{amsfonts}       % blackboard math symbols
\usepackage{nicefrac}       % compact symbols for 1/2, etc.
\usepackage{microtype}      % microtypography
\usepackage{graphicx}
\usepackage{color}
\usepackage{subcaption}

\usepackage{multirow}
\begin{document}

%%%%%%%%% TITLE

\title{Real-time Joint Object Detection and Semantic Segmentation Network for Automated Driving}

\author{
 Ganesh Sistu$^1$, Isabelle Leang$^2$ and Senthil Yogamani$^1$\\
1: Valeo Vision Systems, Ireland\\
2: Valeo Bobigny, France\\
{\tt\small \{ganesh.sistu,isabelle.leang,senthil.yogamani\}@valeo.com}
}

\maketitle

\begin{abstract}

Convolutional Neural Networks (CNN) are successfully used for various visual perception tasks including bounding box object detection, semantic segmentation, optical flow, depth estimation and visual SLAM. Generally these tasks are independently explored and modeled. In this paper, we present a joint multi-task network design for learning object detection and semantic segmentation simultaneously. The main motivation is to achieve real-time performance on a low power embedded SOC by sharing of encoder for both the tasks. We construct an efficient architecture using a small ResNet10 like encoder which is shared for both decoders. Object detection uses YOLO v2 like decoder and semantic segmentation uses FCN8 like decoder. We evaluate the proposed network in two public datasets (KITTI,  Cityscapes) and in our private fisheye camera dataset, and demonstrate that joint network provides the same accuracy as that of separate networks. We further optimize the network to achieve 30 fps for 1280x384 resolution image.
\end{abstract}

%%%%%%%%% BODY TEXT
%%%%%%%%%%%%%%%%%%%%%%%%%%%%%%%%%%%%%%%%%%%%%%%%%%%%%%%%%%%%%%%%%%%%%%%%%%%%%%%%
\section{Introduction} \label{intro}

Convolutional neural networks (CNNs) have became the standard building block for majority of visual perception tasks in autonomous vehicles. Bounding box object detection is one of the first successful applications of CNN for detecting pedestrians and vehicles. Recently semantic segmentation is getting mature \cite{siam2017deep} starting with detection of roadway objects like road, lanes, curb, etc. In spite of rapid progress in computational power of embedded systems and trend in specialized CNN hardware accelerators, real-time performance of semantic segmentation at high accuracy is still challenging. In this paper, we propose a real-time joint network of semantic segmentation and object detection which cover all the critical objects for automated driving. The rest of the paper is structured as follows. Section \ref{mtl4ad} reviews the object detection application in automated driving and provides motivation to solve it using a multi-task network.  Section \ref{proposed} details the experimental setup, discusses the proposed architecture and experimental results. Finally, section \ref{conclusion} summarizes the paper and provides potential future directions.

\section{Multi-task learning in Automated Driving} \label{mtl4ad}

Joint learning of multiple tasks falls under the sub-branch of machine learning called multitask learning. The underlying theory behind joint learning of multiple tasks is that the networks can perform better when trained on multiple tasks as they learn rules of the game faster by leveraging the inter task disciplines. These networks networks are not only better at generalization but also reduces the computational complexity making them very effective for low power embedded systems. Recent progress has shown that CNN can be used for various tasks \cite{horgan2015vision} including moving object detection \cite{siam2018modnet}, depth estimation \cite{kumar2018near} and visual SLAM \cite{milz2018visual}.

Our work is closest to the recent MultiNet \cite{teichmann2018multinet}. We differ by focusing on a smaller network, larger set of classes for the two tasks and more extensive experiments in three datasets. 

\subsection{Important Objects for Automated Driving}

\begin{figure*}[htpb]
\centering
\includegraphics[width=0.6\textwidth]{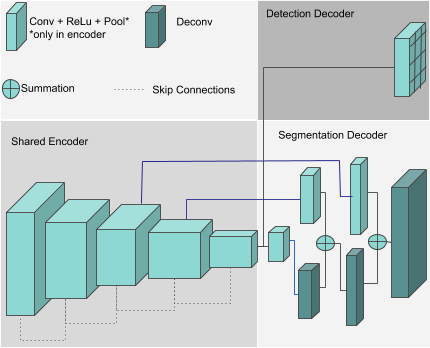}
\caption{Multi-task visual perception network architecture}
\label{fig:multi-task}
\end{figure*}

The popular semantic segmentation automotive datasets are CamVid \cite{brostow2008segmentation} and the more recent Cityscapes \cite{cordts2016cityscapes}. The latter has a size of 5000 annotation frames which is relatively small. The algorithms trained on this dataset do not generalize well to data tested on other cities and with unseen objects like tunnels. To compensate for that, synthetic datasets like Synthia \cite{ros2016synthia} and Virtual KITTI \cite{gaidon2016virtual} were created. There is some literature which demonstrates that a combination produces reasonable results in small datasets. But they are still limited for a commercial deployment of an automated driving system. Hence there is a recent effort to build larger semantic segmentation datasets like Mapillary Vistas dataset \cite{neuhold2017mapillary} and ApolloScape  \cite{huang2018apolloscape}. Mapillary dataset comprises of 25,000 images with 100 classes. ApolloScape dataset comprises of 143,000 images with 50 classes.
% It also offers large variability in terms of weather condition, camera type and geographic coverage. Toronto City is a massive semantic segmentation, mapping and 3D reconstruction dataset covering 712 km$^{2}$ of land, 8439 km of road and around 400,000 buildings. The annotation is completely automated by leveraging Aerial Drone data, HD maps, city maps and LIDARs. It is then manually verified and refined.  

Although semantic segmentation can be used as a unified model for detecting all the objects, there are many issues with this approach. From the application perspective, it is essential to have instances of objects for tracking and path planning. This can be obtained by instance segmentation but it uses bounding box detector as the first step and it is relatively less mature. Annotation for semantic segmentation is time consuming, typically it can take around an hour for annotating a single image which makes it challenging for collecting large datasets. The sample complexity of key bounding box objects like pedestrians and vehicles is much higher compared to road or lanes. Typically legacy datasets exist for these objects using bounding box annotation. It is also relatively easier to do corner case mining for bounding box detectors. Thus there is a need to use both semantic segmentation and object detection models. Additionally, the computational power of embedded system needed for deployment is still a bottleneck. Thus we focus our development on critical object classes. The first important group of classes is roadway objects. They represent the static infrastructure namely road, lane and curb. They cannot be represented by a bounding box and thus need segmentation task. The second group of classes is the dynamic objects which are vehicles, pedestrians and cyclists. This group is particularly important as they are vulnerable road users. 

\subsection{Pros and Cons of MTL}

In this paper, we propose a network architecture with shared encoder which can be jointly learned. The main advantages are increased efficiency, scalability to add more tasks leveraging previous features and better generalization through inductive transfer (learning transferable features for tasks). We address the pros/cons of shared network in more details below. \\
 
\textbf{Pros of shared network:} 

\begin{itemize}
    \item Computational efficiency: The simple and easy to understand intuition behind shared features is improving computational efficiency. Say there are 2 classes and 2 independent networks taking 50\% of processing power individually. If there is a possibility of say 30\% sharing across the 2 networks, each network can re-use the additional 15\% to do a slightly larger network individually. There is plenty of empirical evidence to show that the initial layers of the network are task-independent (oriented Gabor filters) and we should be able to do some level of sharing, more the better. 
    \item Generalization and accuracy: Ignoring computational efficiency, the networks which are jointly learnt tend to generalize better and be more accurate. This is why transfer learning on imagenet is very popular where there network learns very complex classes like differentiating between specific species of dogs. Because the subtle differences between say two species say Labrador or Pomeranian are learnt, they are better at detecting a simpler task of dog detection. Another argument is that there is less possibility of over-fitting for a particular task when they are jointly learnt.
    \item Scalability to more tasks like flow estimation, depth, correspondence, and tracking. Thus a common CNN feature pipeline can be harmonized to be used for various tasks. \\
\end{itemize}

\textbf{Cons of shared network:} 

\begin{itemize}
    \item In case of non-shared network, the algorithms are completely independent. This could make the dataset design, architecture design, tuning, hard negative mining, etc simpler and easier to manage.  
    \item Debugging a shared network (especially when it doesn't work) is harder relatively.
\end{itemize}

\begin{figure*}[htpb]
\centering
\includegraphics[width=\textwidth]{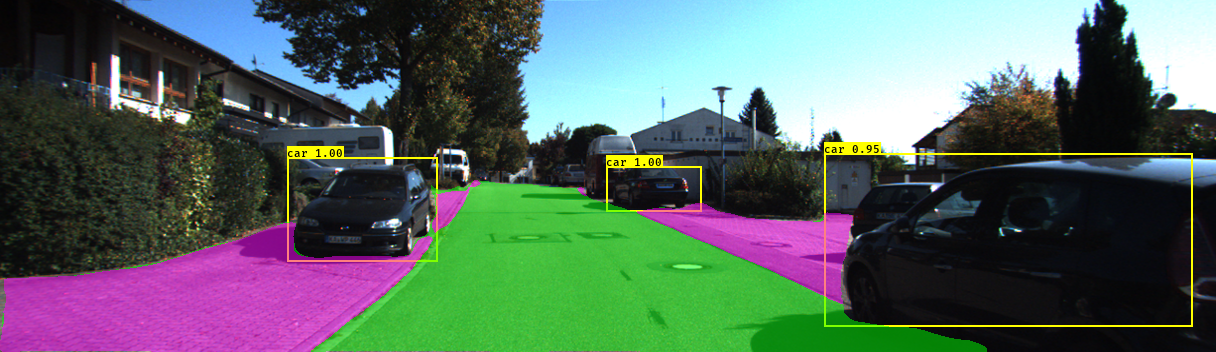}
\caption{Multi-task visual perception network output. Segmentation color coding is green for road and pink for sidewalk}
\label{multi-task}
\end{figure*}

\section{Proposed Algorithm and Results} \label{proposed}

\subsection{Network Architecture}

In this section, we report results on a baseline network design which we plan to improve upon. We propose a jointly learnable shared encoder network architecture provided in the high level block diagram in Figure \ref{fig:multi-task}. We implemented a two task network with 3 segmentation classes (background, road, sidewalk) and 3 object classes (car, person, cyclist). To enable feasibility on a low power embedded system, we used a small encoder namely Resnet10 which is fully shared for the two tasks. FCN8 is used as the decoder for semantic segmentation and YOLO is used as the decoder for object detection. Loss function for semantic segmentation is the cross entropy loss to minimize misclassification. For geometric functions, average precision of object localization is used as error function in the form of squared error loss. We use a weighted sum of individual losses $L = w_{seg}*L_{seg} + w_{det}*L_{det}$ for the two tasks. In case of fisheye cameras which have a large spatially variant distortion, we implemented a lens distortion correction using a polynomial model. 

\subsection{Experiments}

In this section, we explain the experimental setting including the datasets used, training algorithm details, etc and discuss the results.  We trained and evaluated on our internal fisheye dataset comprising of 5000 images and two publicly available datasets, KITTI \cite{geiger2013vision} and Cityscapes \cite{cordts2016cityscapes}. We implemented the different proposed multi-task architecture using Keras \cite{chollet2015keras}. We used pre-trained Resnet10 encoder weights from ImageNet, which is then fine-tuned for the two tasks. The FCN8 upsampling layers are initialized using random weights. We used ADAM optimizer as it provided faster convergence, with a learning rate of 0.0005. Categorical cross-entropy loss and squared errors loss are used as loss function for the optimizer. Mean class IoU (Intersection over Union) and per-class IoU were used as accuracy metrics for semantic segmentation, mean average precision (mAP) and per-class average precision for object detection. %Maximum number of training epochs is set to 30 and early stopping  with a patience of 3 epochs monitoring the gains is added. 
All input images were resized to 1280x384 because of memory requirements needed for multiple tasks. Table \ref{tab:mtl} summarizes the obtained results for STL networks and MTL networks on KITTI, Cityscapes and our internal fisheye datasets. This is intended to provide a baseline accuracy for incorporating more complex multi-task learning techniques.
We compare a segmentation network (STL Seg) and a detection network (STL Det) to a MTL network performing segmentation and detection (MTL, MTL$_{10}$ and MTL$_{100}$).
We tested 3 configurations of the MTL loss, the first one (MTL) uses a simple sum of the segmentation loss and detection loss ($w_{seg}=w_{det}=1$). The two other configurations MTL$_{10}$ and MTL$_{100}$, use a weighted sum of the task losses where the segmentation loss is weighted with a weight $w_{seg}=10$ and $w_{seg}=100$ respectively. This compensates the difference of task loss scaling: the segmentation loss is 10-100 times higher than the detection loss during the training. 
Such a weighting in MTL network improves the performance of the segmentation task for the 3 datasets. Even if the MTL results are slighty lower than the STL results for the segmentation task, this experiment shows that the multi task network has the capacity to learn more by tuning correctly the parameters. Moreover, by keeping almost the same accuracy, we have a drastic gain in memory and computational efficiency. We make use of several standard optimization techniques to further improve runtime and achieve 30 fps on an automotive grade low power SOC. Some examples are (1) Reducing number of channels in each layer, (2) Reducing number of skip connections for memory efficiency and (3) Restricting segmentation decoder to image below horizon line (only for roadway objects).

\begin{table*}[]
\tiny
\centering
\caption{Comparison Study : Single task vs Multi-task}
\begin{tabular}{lllllll}
\hline
Databases                                & Metrics           & STL Seg    & STL Det       & MTL             & MTL$_{10}$      & MTL$_{100}$     \\
\hline
\multirow{4}{*}{KITTI Seg}      & JI background     & 0.9706          &                 & 0.9621          & 0.9663          & 0.9673          \\
                                         & JI road           & 0.8603          &                 & 0.8046          & 0.8418          & 0.8565          \\
                                         & JI sidewalk       & 0.6387          &                 & 0.5045          & 0.5736          & 0.6277          \\
                                         & \textbf{mean IOU} & \textbf{0.8232} & \textbf{}       & \textbf{0.757}  & \textbf{0.7939} & \textbf{0.8172} \\

\hline
\multirow{4}{*}{KITTI Det}         & AP car            &                 & 0.801           & 0.7932          & 0.7746          & 0.7814          \\
                                         & AP person         &                 & 0.469           & 0.5337          & 0.518           & 0.468           \\
                                         & AP cyclist        &                 & 0.5398          & 0.4928          & 0.5107          & 0.5844          \\
                                         & \textbf{mean AP}  & \textbf{}       & \textbf{0.6033} & \textbf{0.6066} & \textbf{0.6011} & \textbf{0.6112} \\
\hline
\hline
\multirow{9}{*}{Cityscapes Seg} & JI road           & 0.9045          &                 & 0.8273          & 0.8497          & 0.8815          \\
                                         & JI sidewalk       & 0.5434          &                 & 0.3658          & 0.4223          & 0.4335          \\
                                         & JI building       & 0.7408          &                 & 0.6363          & 0.6737          & 0.6947          \\
                                         & JI vegetation     & 0.8085          &                 & 0.6949          & 0.7417          & 0.7363          \\
                                         & JI sky            & 0.7544          &                 & 0.6228          & 0.652           & 0.6873          \\
                                         & JI person+rider   & 0.3916          &                 & 0.3225          & 0.3218          & 0.3613          \\
                                         & JI car            & 0.695           &                 & 0.5237          & 0.5918          & 0.6579          \\
                                         & JI bicycle        & 0.3906          &                 & 0.2911          & 0.4123          & 0.3506          \\
                                         & \textbf{mean IOU} & \textbf{0.5971} & \textbf{}       & \textbf{0.4918} & \textbf{0.5213} & \textbf{0.5555} \\
\hline
\multirow{4}{*}{Cityscapes Det}    & AP car            &                 & 0.3691          & 0.411           & 0.398           & 0.3711          \\
                                         & AP person         &                 & 0.1623          & 0.1931          & 0.1694          & 0.1845          \\
                                         & AP bicycle        &                 & 0.1279          & 0.1898          & 0.1422          & 0.1509          \\
                                         & \textbf{mean AP}  & \textbf{}       & \textbf{0.2198} & \textbf{0.2647} & \textbf{0.2365} & \textbf{0.2355} \\
\hline
\hline
\multirow{5}{*}{Our own Seg}      & JI background     & 0.9712          &                 & 0.9346          & 0.9358          & 0.9682          \\
                                         & JI road           & 0.9318          &                 & 0.872           & 0.871           & 0.9254          \\
                                         & JI lane           & 0.6067          &                 & 0.4032          & 0.4426          & 0.594           \\
                                         & JI curb           & 0.506           &                 & 0.1594          & 0.2554          & 0.4507          \\
                                         & \textbf{mean IOU} & \textbf{0.7702} & \textbf{}       & \textbf{0.606}  & \textbf{0.6419} & \textbf{0.7527} \\
\hline
\multirow{3}{*}{Our own Det}         & AP car            &                 & 0.5556          & 0.564           & 0.554           & 0.5571          \\
                                         & AP person         &                 & 0.3275          & 0.3639          & 0.2971          & 0.3609          \\
                                         & \textbf{mean AP}  & \textbf{}       & \textbf{0.4415} & \textbf{0.4639} & \textbf{0.4255} & \textbf{0.459} \\
\hline
\end{tabular}
\label{tab:mtl}
\end{table*}

% \begin{figure}[!t]
% \centering
% \includegraphics[width=0.5\textwidth]{undistort}
% \caption{Visualization of rectilinear correction to remove fisheye distortion. Top image is original and bottom image is rectified.}
% \label{fig:undistort}
% \end{figure}

\section{Conclusion} \label{conclusion}
In this paper, we discussed the application of multi-task learning in an automated driving setting for joint semantic segmentation and object detection tasks. Firstly, we motivated the need to do both tasks instead of just semantic segmentation. Then we discussed the pros and cons of using a multi-task approach. We constructed an efficient joint network by careful choice of encoders and decoders and further optimize it to achieve 30 fps on a low-power embedded system. We shared experimental results on three datasets demonstrating the efficiency of joint network. In future work, we plan to explore addition of visual perception tasks like depth estimation, flow estimation and Visual SLAM.

%\vfill

% \section*{Acknowledgement}

% The authors would like to thank their employer for the opportunity to work on fundamental research. We would also like to thank Niall Ryan (Valeo), Lucie Yahiaoui (Valeo) and B Ravi Kiran (Akka) for reviewing the paper and providing feedback.

% \bibliographystyle{ieee}
% \bibliography{egbib}

{
\medskip
\small
\bibliographystyle{plain}
\bibliography{egbib}
}

\end{document}